%% file: main.tex
\documentclass[letterpaper, 10pt, conference]{ieeeconf}

\IEEEoverridecommandlockouts
\usepackage{cite}
\usepackage{amsmath,amssymb,amsfonts}
\usepackage{algorithm}
\usepackage{algorithmic}
\usepackage{multirow}
\usepackage{stfloats, graphicx}
\usepackage{xcolor}
\usepackage{textcomp}
\usepackage{svg}
\usepackage[outdir=./]{epstopdf}
\usepackage{generic}

\newtheorem{lemma}{Lemma}
\newtheorem{theorem}{Theorem}
\newenvironment{proofof}{\noindent {\em Proof of }}{\hfill \hspace*{1pt}
\hfill $\blacksquare$}

\title{\LARGE \bf A Passivity Preserving H-infinity Synthesis Technique for Robot Control}

\author{Daniel Larby, Fulvio Forni
	\thanks{This work was supported by the Engineering and Physical Sciences Research Council  [EP/T517847/1]; and by CMR Surgical. For the purpose of open access, the author has applied a Creative Commons Attribution (CC BY) licence to any Author Accepted Manuscript version arising. D Larby and F. Forni are with the Department of Engineering, University of Cambridge, CB2 1PZ, UK.
		{\tt\footnotesize dl564.cam.ac.uk|f.forni@eng.cam.ac.uk}
	}
}

\begin{document}

%\IEEEpubid{\makebox[\columnwidth]{978-1-5386-5541-2/18/\$31.00~\copyright2018 IEEE \hfill} \hspace{\columnsep}\makebox[\columnwidth]{ }}
\maketitle
%\IEEEpubidadjcol

\input{content/0-Abstract}
\input{content/1-Introduction}

\input{content/14-Section2}

\input{content/15-Section3}

\input{content/16-Section4}
\input{content/13-Conclusions}

\bibliographystyle{plain}
\bibliography{./../../../zotero/library}

\end{document}

%% file: content/0-Abstract.tex
\begin{abstract}
Most impedance control schemes in robotics implement a desired passive impedance,
allowing for stable interaction between the controlled robot and the environment.
However, there is little guidance on the selection of the desired impedance. In
general, finding the best stiffness and damping parameters is a challenging task.
This paper contributes to this problem by connecting impedance control to robust control,
with the goal of shaping the robot performances via feedback.
We provide a method based on linear matrix inequalities with sparsity constraints to
derive impedance controllers that satisfy a $\mathcal{H}_\infty$ performance criterion.
Our controller guarantees passivity of the controlled robot and local performances 
near key poses.
\end{abstract}

%% file: content/1-Introduction.tex
\section{Introduction}

There are many considerations when  designing a controller for a robotic manipulator covering a wide range of tasks such as pick-and-place operations, cooperative robotics, teleoperated robots and robotic surgery. 
Any successful control approach must guarantee stability and must remain stable when interacting with the environment.
A good approach would also shape the robot performance.
However, doing so analytically is made difficult by the nonlinear nature of the dynamics of robot manipulators.

Passivity-based control (PBC) is a powerful technique widely used in control design for robotics \cite{Chopra2022, Ortega1998, Hatanaka2015}. PBC provides strong stability guarantees, via straightforward energy considerations. %
The most widely used controller in robotics, the joint-space PD control with gravity compensation, is
a simple and robust passivity-based controller \cite{Takegaki1981}. Its energy shaping and damping injection features guarantee passivity of the controlled robot. Passivity-based control design techniques have also been adopted for parameter estimation \cite{Slotine1987}, remote-teleoperation \cite{Anderson1989}, 
and trajectory tracking via passive-aware adaptive estimation of the parameters of the system \cite{Slotine1988}.

Most impedance control schemes \cite{Hogan1984, Whitney1985} implement a passive impedance defined by a desired stiffness, damping, and inertia.
An impedance controller implementing a desired stiffness and damping (without modifying the inertia) by torque/force controlled joints is a form of PBC energy-shaping and damping injection.
Thus, it shares the same stability properties. At the same time, impedance control opens the way to the characterization and control of the robot performances, via loop shaping and similar considerations \cite{Colgate1988}.
However, in practice, there is little guidance on the selection of desired impedances. In fact, the recent review  \cite{Song2019} concludes that there is a need to establish more systematic guidance or methodology on how to specify the impedance parameters, such as desired stiffness, damping and inertia.

In the recent review paper \cite{Hogan2022}, Hogan pointed out that a reasonable choice of desired stiffness and damping matrices is diagonal, based on criteria of simplicity and effectiveness. We describe in Section \ref{background_ES} how this method of impedance control corresponds to energy shaping and damping injection.
In fact, drawing on the mechanical analogy of the controller as a virtual mechanical system \cite{Pratt2001}, this method corresponds to the interconnection of the robot with a system of Cartesian (or otherwise 1-dimensional) virtual springs and dampers.
In \cite{VanDerSchaft2020} Van der Schaft also remarks that optimising among achievable impedances for a task is an important problem that needs to be studies in more depth. Some existing approaches for selecting impedances involve machine learning for variable impedance control \cite{Abu-Dakka2020, Bogdanovic2020}, but by varying the stiffness, passivity of the controller is lost. Methods to preserve passivity exist, such as tank-based variable impedance control \cite{Ferraguti2013}, but using this approach requires a careful consideration to ensure that ``energy-tanks'' do not deplete \cite{Schindlbeck2015}, hindering performance.

This paper contributes to the challenge of impedance selection / optimization, by casting the problem into a robust control problem. Our approach takes advantage of the robot linearization in operation space. In this setting, a static-state feedback $\mathcal{H}_\infty$ synthesis can be performed to systematically select feedback gains. Our design is based on linear matrix inequalities and forces the feedback gains to be diagonal and positive, which corresponds to synthesising a set of diagonal passive impedances, as suggested in \cite{Hogan2022}. This provides a systematic approach to selecting impedance parameters for a common class of impedance controllers, based on criteria of optimality with respect to $\mathcal{H}_\infty$ metrics.

Section \ref{background_ES} introduces the framework of energy shaping and damping injection as the interconnection of the robot with virtual spring and dampers. Working in operation space, this  allows for intuitive considerations on the shaping of the robot potential energy. It also allows to draw a simple comparison with impedance control. Section \ref{background_LMI} discusses operation space impedance control, building the connection between closed-loop impedance shaping and robust control synthesis. The main contribution of the paper is in Section \ref{solution}, which provides the design method based on linear matrix inequalities to synthesize automatically stiffness and damping parameters of our impedance controller. The synthesis is driven by $\mathcal{H}_\infty$ gain criteria. Our approach is illustrated on a $2$-link and on a $7$-link manipulator in Section \ref{examples}. Conclusions follow.

%% file: content/14-Section2.tex
\section{Energy shaping and damping injection} \label{background_ES}

Energy shaping and damping injection methods are well known in robotics and the control of electromechanical systems in general \cite{Willems1972, Ortega2001}.
	The core idea is that a \textit{passive} system may store energy, route it between ports or dissipate it, but it may not produce energy. When two passive systems are connected together by compatible ports, then the resulting combined system is passive. 
The shaping of the energy function changes the equilibria of the system, a feature that can be used to control the robot to a desired position.
The damping removes energy, affecting the speed of convergence. In general, the design of an energy shaping and damping injection controller to stabilize a nonlinear system at a particular equilibrium could require solving a partial differential equation \cite{Ortega2001}.
However, for fully actuated mechanical systems we can take a shortcut. 
Consider a robot system 
$$
M(q)\ddot{q} + C(q,\dot{q})\dot{q} + g(q) = u
$$
with energy 
$H(q, \dot q) = \frac{1}{2}\dot q^T M(q)\dot q + U(q)$, where $M(q) = M(q)^T \succ 0$ is the generalized inertia matrix and $U(q)$ the gravitational potential energy. 
The gravity force/torque vector is obtained by taking the partial derivative of the gravitational potential energy: $g(q) = \frac{\partial U}{\partial q}(q)$. 
The simplest way to shape the energy function is to select 
$$
H_d(q, \dot q) = \frac{1}{2}\dot q^T M(q)\dot q + \frac{1}{2}  (q-\bar q)^T \mathrm{diag}(k)(q-\bar q),
$$ 
where the elements of the vector $k$ are all strictly positive (stiffness terms). This can be achieved with the controller 
$$
u = \mathrm{diag}(k) (\bar q - q) +  \frac{\partial U(q)}{\partial q}  - \mathrm{diag}(b) \dot{q},
$$
where the vector $b$ has strictly positive elements. The term $- \mathrm{diag}(b) \dot{q}$ ensures dissipation and guarantees that the system trajectories
converge asymptotically to the minimum of this energy function at $q = \bar q$ and $\dot q = 0$.
This is of course joint-space control with gravity compensation. Its simplicity and mechanical interpretability as springs and damper at the robot's joints makes it one of the most used approaches in industrial robotics. 

\begin{figure}[h]
	\centering
	\includegraphics[width=.76\columnwidth]{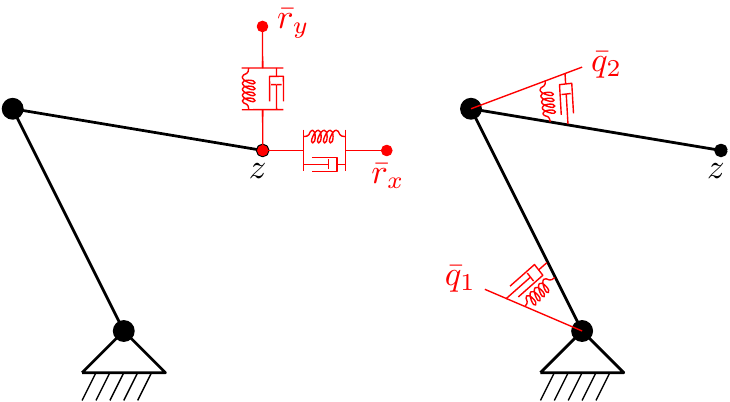} 
	\vspace{-1mm}
	\caption{A comparison of PD control in operation-space (left) and joint-space (right). The actions of the controllers are represented in red.}
	\label{fig:twolink}
\end{figure}

Energy shaping and damping injection is not confined to the joint space. In fact, moving away from joint space allows for advanced design of the robot's response to external perturbations and avoids the computation of $\bar q$ via inverse kinematics. The idea is to shape the energy of the robot in a space that is closer to the control objective. For position control of the end-effector of a planar manipulator, for example, we just need to attach a pair of springs on Cartesian axes between the end-effector position, $z_c$, and the reference position, $r$. This is visualized in Figure \ref{fig:twolink}. 
Using the forward kinematics $z_c=h_c(q)$,  the corresponding desired energy function is 
\begin{equation}
	\label{eq:energy_os}
	H_d(q, \dot q) = \frac{1}{2}\dot q^T M(q)\dot q + 
	\frac{1}{2}  (h_c(q)- r)^T \mathrm{diag}(k)(h_c(q)-r).
\end{equation}
This can be achieved with the controller
\begin{equation}
	\label{eq:control_law_u}
	u = J_c(q)^T F_c + \frac{\partial U(q)}{\partial q}
\end{equation}
 and 
\begin{equation}
	\label{eq:control_law_Fc}
	F_{c} = \mathrm{diag}(k) (r-h_c(q)) - \mathrm{diag}(b) J_c(q)\dot q,
\end{equation}
where $J_c(q) = \frac{\partial h_c(q)}{\partial q}$.

The control action $u$ emulates the effect of the force $F_c$ at the end-effector,
realizing a proportional and derivative control action in the coordinates of the end-effector. 
This approach offers several advantages. The potential energy is a function of end-effector displacement. 
This implies that the response of the robot to external perturbations 
is independent of the robot's configuration. Furthermore, 
there is no need to implement an inverse kinematics to drive the robot to a desired position $r$. 
This is particularly important for redundant robots, where several configurations can attain the same end-effector position.
In fact, for redundant robots, we can control the end-effector position without forcing high-stiffness on each joint.

In this paper we will focus on energy shaping and damping injection controllers that can be represented as 
virtual springs and dampers attached to various points of the robots. These are implemented by the
control law \eqref{eq:control_law_u}, \eqref{eq:control_law_Fc}.
The combination of several virtual elements allows to a straightforward tuning of the energy landscape while preserving passivity of
the robot, for stable interactions with the environment. 
The placement of virtual springs and dampers is dictated by the features of the task and by considerations 
on the robot's dynamics. Once a placement is selected, the choice of the parameter vectors $k$ and $b$ will affect the performance of the system. 
Therefore, it is natural to ask if there is a systematic framework for the selection of $k$ and $b$ taking into account performance considerations.
This will be discussed in the next section, from the perspectives of impedance control and loop shaping.

%% file: content/15-Section3.tex
\section{Impedance control} \label{background_LMI}
A systematic framework for the selection of the control parameters $k$ and $b$
requires to main factors: (i) a way to measure the performance of the robot and (ii) a numerical algorithm
to optimize such performance. For the first objective, we will extend the robot equations with
auxiliary input/output variables as follows:
\begin{equation}
	\label{eq:dynamics_extended}
	\begin{split}
		M(q)\ddot{q} + C(q,\dot{q})\dot{q} + g(q) &= u_c + J_d(q)^T F_d \\
		z_d &= h_d(q) \,.
	\end{split}
\end{equation}
The \emph{control} pair $(u_c,(q,\dot{q}))$ will be used to design a controller of the form \eqref{eq:control_law_u}, \eqref{eq:control_law_Fc}.
This corresponds to selecting $k$ and $b$ once the structure of the controller has been decided (i.e. the placement of springs and dampers).
The  \emph{performance} pair $(F_d,z_d)$ captures the effect that a perturbing 
exogenous force $F_d$ has on the displacement $z_d$.
High performance position control, for example, would require insensitivity of the displacement $z_d$ to the perturbing force $F_d$. 
This corresponds to enforcing a \emph{low gain} between $F_d$ and $z_d$. In contrast, a compliant robot would allow for large
displacements $z_d$ for small perturbations $F_d$. 
In this setting, the selection of the parameters $k$ and $b$ can be guided by optimization. For a controller of the form \eqref{eq:control_law_u}, \eqref{eq:control_law_Fc}, the goal is to find the best selection of $k$ and $b$ that optimize some performance metric on $(F_d,z_d)$.
\begin{figure}[h]
	\centering
	\includegraphics[width=.45\columnwidth]{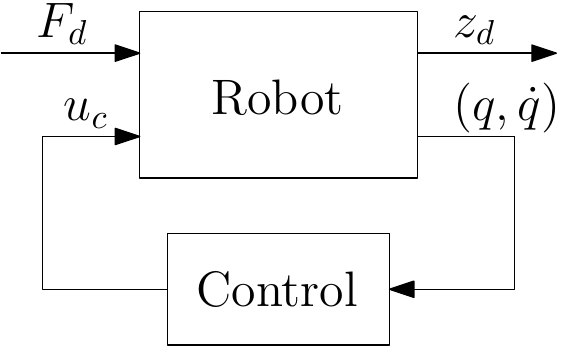} 
	\vspace{-1mm}
	\caption{Feedback control loop to optimize performances on $(F_d,z_d)$.}
	\label{fig:control}
\end{figure}

The control task is illustrated in Figure \ref{fig:control}. We close the loop on the control pair $(u_c,(q,\dot{q}))$ 
with a controller of the form \eqref{eq:control_law_u}, \eqref{eq:control_law_Fc} with the goal of shaping the behaviour captured by the performance pair $(F_d,z_d)$. This setting finds contacts with impedance control \cite{Hogan1984} and classical robust control synthesis \cite{Zhou1996}.
Indeed, impedance control recognizes that positions and forces cannot be controlled independently and the best we can do is to shape the relationship
between these quantities. Likewise, in the linear setting, robust control provides methods to find a controller that minimize the gain between $F_d$ and $z_d$.

In what follows we will assume that $z_c = h_c(q)$ in \eqref{eq:control_law_u}, \eqref{eq:control_law_Fc} is given. 
That is, we start from a predefined placement of virtual springs and dampers attached to various points of the robots.
This corresponds to the particular shaping of the desired energy given by \eqref{eq:energy_os}. We assume that $h_c$ is a differentiable invertible function 
within the region of interest. This will allow us to derive an equivalent representation of the robot in the coordinates $z_c$, 
which suits linearization methods for state-feedback design. We will look into controllers of the form \eqref{eq:control_law_u}, \eqref{eq:control_law_Fc} 
and we will design the control parameters $k$ and $b$ to shape the relationship between $F_d$ and $z_d$, following the approach of impedance control.

\begin{figure}[h]
	\centering
	\includegraphics[width=.9\columnwidth]{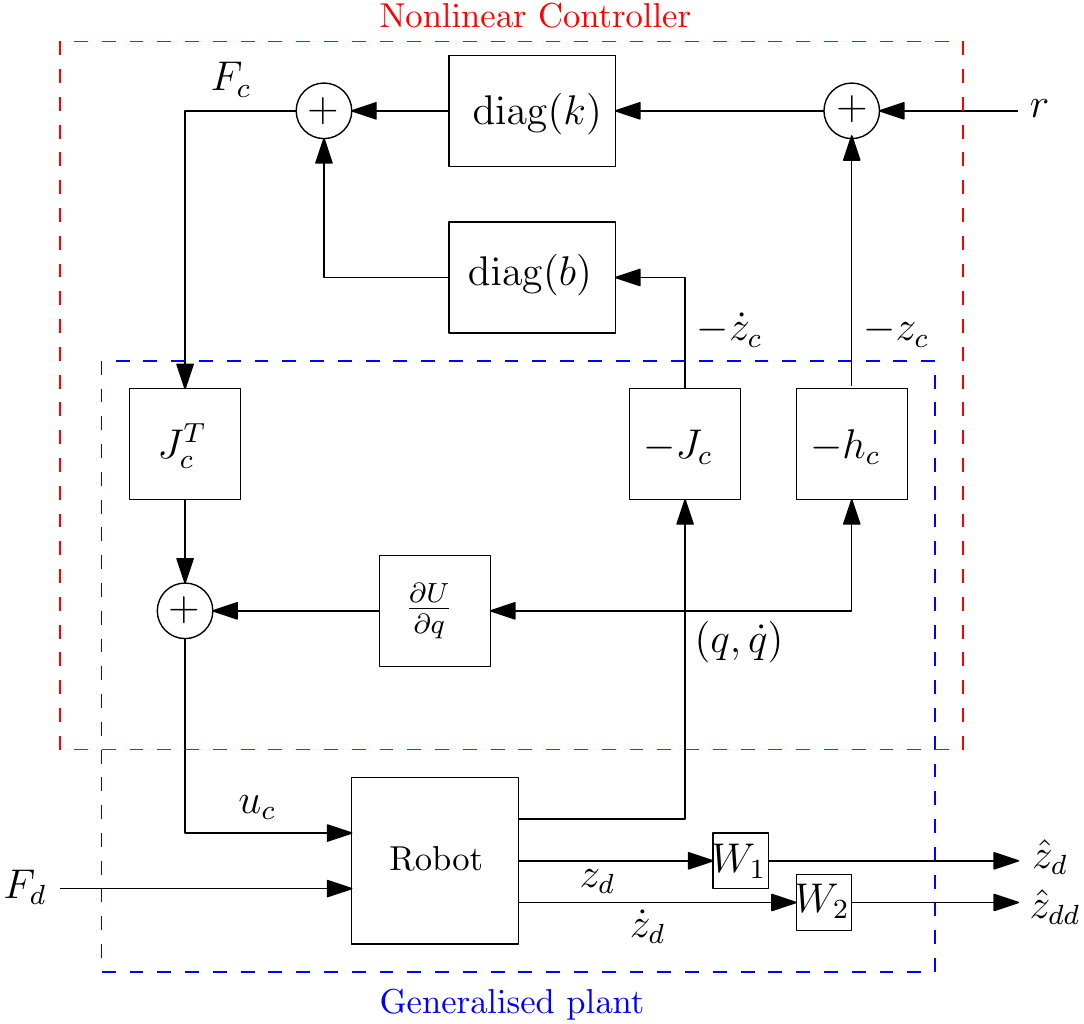} 
	\caption{The overall structure of the controller connects the energy shaping and damping injection approach with impedance control / 
		$\mathcal{H}_\infty$ synthesis of the controller parameters $k$ and $b$.}
	\label{fig:hinf_setup_FF}
\end{figure}

The actual synthesis of the control parameters will be performed via $\mathcal{H}_\infty$ synthesis \cite{Zhou1996}.
Specifically, we will use linear matrix inequalities \cite{Boyd1994} to derive a state-feedback controller that optimizes \emph{locally}
the performances captured by $(F_d,z_d)$, measured via $\mathcal{H}_\infty$ metric.
We will allow for the optimization of weighted performances represented by 
\begin{equation}
	\label{eq:weighted_variables}
	\hat{z}_{d} = W_1 z_d \qquad \hat{z}_{dd} = W_2 \dot{z}_d
\end{equation}
The role of $W_1$ and $W_2$ is to scale non-uniformly the components of $z_d$ and of its derivative.
For simplicity of the exposition they are constant, real matrices. 
However, the design can be easily extended to frequency dependant weights.
We will show how our conditions based on linear matrix inequalities lead to control parameters $k$ and $b$ that 
preserve the passivity of the robot.
The overall structure of the controller is summarized in Figure \ref{fig:hinf_setup_FF}.

\section{Passivity and local $\mathcal{L}_2$ gains} \label{solution}

\subsection{The robot linearization in operative space} \label{OS_linearisation}

The selection of the control parameters $k$ and $b$ to optimize local performances can be derived via linear matrix inequalities (LMIs). 
The first step is to derive the robot linearization in operation space. 
We will then determine a state feedback $K$ via LMIs, which will lead to the control law \eqref{eq:control_law_Fc} of the form 
$$F_{c} = -K \left[\begin{array}{c}  h_c(q)-r \\ J_c(q) \dot q \end{array}\right] \, .$$

Representing the dynamics in operation space corresponds to making a different choice of general coordinates when deriving the robot dynamics \cite{Khatib1987}. As a result, the operation space dynamics must have dimension $N$, the number of degrees of freedom of the robot. In practice, 
it is difficult to derive this dynamics directly from Lagrangian methods for arbitrary operation space variables. It is easier to draw an equivalence between the terms of the joint space dynamics, $M(q)$, $C(q, \dot q)$, and $g(q)$, and their operation-space equivalents \cite{Khatib1987}.
Differentiating $\dot z_c$ gives the acceleration
$$
\ddot z_c = \frac{d}{dt} \left(J_c(q)\dot q\right) = \dot J_c(q)\dot q + J_c(q)\ddot q 
$$
Combining this with our control law \eqref{eq:control_law_u} and dynamics \eqref{eq:dynamics_extended}, assuming a nonsingular and therefore invertible Jacobian gives us the operation space dynamics:
\begin{equation} 
	\begin{split}
		&\Delta(q) \ddot z_c + \mu(q, \dot q) \dot z_c = F_c + J_c(q)^{-T}J_d(q)^TF_d \\[5pt] %
		\Delta(q) &= J_c(q)^{-T}M(q) J_c(q)^{-1} \\[5pt]
		\mu(q, \dot q) &= \Delta(q)\dot J_c(q) + J_c(q)^{-T}C(q, \dot q)J_c(q)^{-1}
	\end{split} \nonumber
\end{equation}

For completeness, arguments $q$ and $\dot q$ should be written as functions of $z_c$, by the inverse kinematics $z_c=h_c(q)$, and we should operate in the region of invertibility of $h_c(q)$. However, in practice there are good reasons to keep the arguments in joint variables.
The inverse kinematics are non-unique,
so characterizing and finding inverse kinematics functions is in general hard, and there is no need to. Instead we can compute $\Delta$ directly from a pose $q^*$, without explicitly defining an inverse kinematics $q^* = h_c^{-1}(z^*)$, so long as the forward kinematic is satisfied and $z^* = h_c(q^*)$.

Implicitly linearizing at some equilibrium, that is, $\ddot z_c = 0$, $\dot z_c = 0$, $q = q^*$, $z_c = z^*$, $F_c = 0$, $F_d = 0$, and using $\tilde{z}_c \simeq z_c-z^*$ to denote small displacement variables, 
we get the linearized operation space dynamics
\begin{equation}
	\label{ddz_linearisation}
	\ddot{\tilde{z}}_c
	= \Delta(q^*)^{-1} 
	\left( 
	F_c + J_c(q^*)^{-T} J_d(q^*)^{T}F_d
	\right) \, .
\end{equation}
Using the state variable $x = [z_c-z^*; \dot{z}]$ and the output variable $y =  [\hat{z}_d-W_1 h_d(q^*), \hat{z}_{dd}]$ from \eqref{eq:weighted_variables}, \eqref{ddz_linearisation} leads to the linearized state-space system
\begin{subequations}
	\label{linearisation}
	\begin{align}
		\dot x(t) &= Ax(t) + B_1F_c(t) + B_2 F_d(t) \label{LMIdynamics}\\
		y(t) &= C x(t)  \label{subeq:LMIy} \\
		F_c(t) &= -Kx(t) \label{subeq:LMI_Fc}
	\end{align}
\end{subequations}
with matrices 
\begin{align*}
	A &\!=\! \begin{bmatrix}
		0 & I \\
		0 & 0
	\end{bmatrix}\! ,
	\!\!\!&\!\!\!	
	B_2 &\!=\! \begin{bmatrix}
		0 \\
		\Delta(q^*)^{-1} J_c(q^*)^{-T}J_d(q^*)^T
	\end{bmatrix} \\
	B_1 &\!=\! \begin{bmatrix}
		0 \\
		\Delta(q^*)^{-1}
	\end{bmatrix}\! ,
	\!\!\!&\!\!\!	
	C &\!=\! \begin{bmatrix}
		W_1 J_d(q^*\!) J_c(q^*\!)^{-\!1}\!\!\!\!\!&\!\!\!\!\! 0 \\
		0 \!\!\!\!\!&\!\!\!\!\! W_2 J_d(q^*\!) J_c(q^*\!)^{-\!1} \\
	\end{bmatrix}\!.
\end{align*}

\subsection{Local $\mathcal{H}_\infty$ synthesis}
\label{sec:H_inf_syn}

Consider a selection of poses $q_i^*$, the associated linearization $\mathcal{P}_i$ given by \eqref{linearisation},
and the matrix 
\begin{equation}
	\label{eq:basic_LMI_matrix}
	\begin{split}
		&\overline{M}(\mathcal{P}_i,\gamma,Q,L) := \\
		&		\left[
		\begin{array}{ccc} 
			QA_i^T - L^TB_{1,i}^T + A_iQ - B_{1,i}L & B_{2,i} & QC_i^T \\
			B_{2,i}^T & -\gamma I & 0 \\
			C_i Q & 0 & -\gamma I 
		\end{array}
		\right]
	\end{split}
\end{equation}
where the subscript $i$ is used to denote the matrices associated with the linearization $\mathcal{P}_i$.
$Q$ and $L$ are matrices of opportune dimensions, and $\gamma > 0$ is a scalar.

Any solution to the LMI problem below returns a uniform state-feedback controller \eqref{subeq:LMI_Fc} given by $$K = LQ^{-1}$$
that guarantees $\mathcal{L}_2$ gain
less than $\gamma>0$ from the input $F_d$ to the output $y$, for each linearization $\mathcal{P}_i$. 

\begin{figure}[h!]
	\noindent{\begin{center}\textbf{LMI 1: $\mathcal{H}_\infty$ synthesis for multiple linearizations}\end{center}}
	\vspace*{-0.1cm}
	\hrule
	\begin{align*}
		\text{Find} &\quad Q=Q^T \in \mathbb{R}^{2N\times 2N}, L\in \mathbb{R}^{N\times 2N} \\
		\text{s.t.}  & \quad Q \succ 0 \\ 
		& \quad \overline{M}(\mathcal{P}_i, \gamma, Q, L)  \prec 0 \quad  \mbox{for each linearization } \mathcal{P}_i
	\end{align*}	
	\hrule \vspace{0.4cm}
\end{figure}

The LMI formulation above is well-known \cite[Section 7]{Boyd1994} and is based on the repeated use of \eqref{eq:basic_LMI_matrix} 
for each linearization $\mathcal{P}_i$. The LMI problem is efficiently solved by packages for convex optimization like CVX \cite{Grant2008}.
Feasibility of this LMI problem for small  $\gamma$ guarantees that exogenous perturbing forces $F_d$ induce small perturbations 
(in the $\mathcal{L}_2$ sense) on positions and velocities of the robot, locally around each equilibrium $q_i^*$.
Indeed, a small $\gamma$ guarantees good performance on the controlled position locally near each equilibrium.

\subsection{$\mathcal{H}_\infty$ synthesis for energy shaping and damping injection}
\label{sec:sparse_H_inf_syn}
The method discussed in Section \ref{sec:H_inf_syn} synthesizes state-feedback gains with guaranteed performances. However, the derived state-feedback controller does not match the energy shaping and damping injection structure of Section \ref{background_ES}. In particular, for the synthesis of stiffness and damping parameters vectors $k$ and $b$ we need 
\begin{itemize}
	\item a state-feedback $K = [K_1 \, K_2]$ with block diagonal matrices $K_1 = \mathrm{diag}(k)$ and $K_2 = \mathrm{diag}(b)$;
	\item all the elements of $K_1$ and $K_2$ must be strictly positive, to guarantee that the control action \eqref{eq:control_law_u}, \eqref{eq:control_law_Fc} leads to a passive controlled robot.
\end{itemize}
To achieve this, we propose below a modified LMI problem with additional constraints to ensure that $K_1$ and $K_2$ are positive definite diagonal matrices.
These diagonality constraints lead to a conservative LMI formulation, but any solution leads to a controller that guarantees passivity and local performances in terms of $\mathcal{L}_2$ gain $\gamma$, as clarified in Theorem \ref{thm:main_result}.

\begin{figure}[h!]
	\noindent{\begin{center}\textbf{LMI 2: sparse $\mathcal{H}_\infty$ synthesis and passivity}\end{center}}
	\vspace*{-0.1cm}
	\hrule
	\begin{align*}
		\text{Find} &\quad Q_{11},  Q_{12}, Q_{22}, L_1, L_2 \in \mathbb{R}^{N\times N} \\
		\text{s.t.}  & \quad Q_{11}, Q_{12}, Q_{22}, L_1, L_2 \mbox{ are all diagonal matrices} \\
		& \quad  Q  = \begin{bmatrix}
			Q_{11} & Q_{12} \\
			Q_{12} & Q_{22}
		\end{bmatrix} \succ 0 \,,\ Q_{12} \prec 0\\
		& \quad L = [\,L_1 \ L_2\,] \,,\ L_{1} \succ 0 \,,\ L_{2} \succ 0 \\
		& \quad \overline{M}(\mathcal{P}_i, \gamma, Q, L)  \prec 0 \quad  \mbox{for each linearization } \mathcal{P}_i
	\end{align*}	
	\hrule \vspace{0.4cm}
\end{figure}

\begin{lemma}
	\label{lem:diag_positivity}
	For any solution to \textbf{LMI 2}, take $K_1 \in \mathbb{R}^{N\times N}$ and $K_2 \in \mathbb{R}^{N\times N}$ such that $[K_1 \ K_2] = LQ^{-1}$.
	
	Then, $K_1$ and $K_2$ are diagonal positive definite matrices. 
\end{lemma}
\vspace{2mm}

\begin{theorem}
	\label{thm:main_result}
	Let $q_i^*$ be the selection of poses with associated linearization $\mathcal{P}_i$ in \textbf{LMI 2}.
	For any solution to \textbf{LMI 2}, take $k$ and $b$ in \eqref{eq:control_law_Fc} such that 
	$[\mathrm{diag}(k) \ \mathrm{diag}(b)] = LQ^{-1}$.
	
	Then, the controlled robot \eqref{eq:control_law_u}, \eqref{eq:control_law_Fc}, \eqref{eq:dynamics_extended}
	\begin{itemize}
		\item is passive at any port $(F_e,\dot{z}_e)$;
		\item satisfies the local $\mathcal{L}_2$ gain 
		$$\left\| \begin{bmatrix} W_1(z_d-h_d(q_i^*)) \\ W_2 \dot{z_d} \end{bmatrix} \right\|_2 \leq \gamma \|F_d\|_2$$
		for initial condition and reference $z_c(0) = r = h_c(q^*_i)$, for any sufficiently small $F_d$.
	\end{itemize}
\end{theorem}
\vspace{2mm}

Theorem \ref{thm:main_result} clarifies that the control action obtained from \textbf{LMI 2} guarantees passivity of the 
controlled robot at any port $(F_e,\dot{z}_e)$, where $F_e$ is a vector of forces (or torques) and $\dot{z}_e$ is the corresponding
vector of co-located velocities (or angular velocities). This is crucial for stable interaction. 
In fact if $J_c$ is full rank the controller guarantees strict output passivity.
At the same time, the controller guarantees
local performances, as long as the perturbing force $F_d$ is small enough to not invalidate the linear approximation \eqref{linearisation}.

\subsection{Proofs}

\begin{proofof}
	\emph{Lemma \ref{lem:diag_positivity}}.
	Define
	\begin{equation}
		\label{eq:P}
		P = \begin{bmatrix}
			P_{11} & P_{12} \\
			P_{12}^T & P_{22} 
		\end{bmatrix} = Q^{-1} .
	\end{equation} 
	Then, 
	\begin{equation}
		\label{eq:K1K2}
		[\,K_1 \ K_2\,] 
		= [\,L_1 \ L_2\,]   P 
		= \begin{bmatrix} 
			L_1 P_{11} \!+\! L_2 P_{12}^T \!&\! L_1P_{12} \!+\!  L_2 P_{22} 
		\end{bmatrix}\!.
	\end{equation}
	Note also that, by block inversion, $Q^{-1} = R_1 R_2$ where 
	$$
	R_1 =	
	\begin{bmatrix}
		(Q_{11}-Q_{12}Q_{22}^{-1}Q_{12}^T)^{-1} & 0 \\
		0 & (Q_{22}-Q_{12}^TQ_{11}^{-1}Q_{12})^{-1}
	\end{bmatrix} 
	$$		
	$$
	R_2 =
	\begin{bmatrix}
		I & -Q_{12}Q_{22}^{-1} \\
		-Q_{12}^T Q_{11}^{-1} & I
	\end{bmatrix} \, .
	$$
	
	With these definitions we can now prove that $K_1$ and $K_2$ are diagonal matrices. 
	Recall that if a matrix is diagonal, it remains so after inversion and transposition. Furthermore, product and sum of diagonal matrices lead to diagonal matrices. Since $P = R_1R_2$, this implies that $P_{11}$, $P_{12}$, $P_{22}$ are diagonal matrices. Thus, from \eqref{eq:K1K2}, also $K_1$ and $K_2$ are diagonal matrices.
	
	We can also prove that $K_1$ and $K_2$ are positive definite. 
	For instance, recall that
	$Q \succ 0 $ implies $P \succ 0$, therefore $Q_{11} \succ 0$, $Q_{22} \succ 0$, $P_{11} \succ 0$, and $P_{22} \succ 0$.
	Furthermore, recall that inversion preserve positivity, and that the product and the sum of two positive diagonal matrices are positive and diagonal.
	Thus, using again $P= R_1R_2$, we have
	$$
	P_{12} = \underbrace{P_{11}}_{\succ 0} (\underbrace{-Q_{12}}_{\succ 0})\underbrace{Q_{22}^{-1}}_{\succ 0}  \succ 0 ,
	$$
	where we have used the assumption  $Q_{12} \prec 0$ in \textbf{LMI 2}. 
	Finally, using \eqref{eq:K1K2} with $L_1 \succ 0$ and $L_2 \succ 0$ from \textbf{LMI 2}, 
	we can conclude that $K_1$ and $K_2$ are positive definite.
\end{proofof}

\newpage

\begin{proofof}
	\emph{Theorem \ref{thm:main_result}}.

We first prove passivity.
Consider the controller \eqref{eq:control_law_u}, \eqref{eq:control_law_Fc}.
By Lemma \ref{lem:diag_positivity}, $\mathrm{diag}(k)$ and $\mathrm{diag}(b)$ are positive definite matrices. 
For the robot equations extended with a generic additional port
\begin{equation}
	\label{eq:dynamics_extended2}
	\begin{split}
		M(q)\ddot{q} + C(q,\dot{q})\dot{q} + g(q) &= u_c + J_e(q)^T F_e \\
		z_e &= h_e(q) \,.
	\end{split}
\end{equation}
where $J_e(q) = \frac{\partial h_e(q)}{ \partial q}$, consider the energy function \eqref{eq:energy_os}.
This is a well defined energy function since $\mathrm{diag}(k) \succ 0$.  
Then,
$$
\dot{H}_d(q,\dot{q}) = - \dot{q}^T J_c(q)^T \mathrm{diag}(b) J_c(q)\dot q + \dot{z}^T_e F_e \leq \dot{z}^T_e F_e
$$
where the inequality follows from $\mathrm{diag}(b) \succ 0$. This proves passivity.

We now show the local gain inequality. Recall that $z_c = h_c(q)$, take $z^*_i = h_c(q_i^*)$, and consider the local storage function 
$$
V(z_c-z^*_i, \dot{z}) = 
\begin{bmatrix}
	z_c-z^*_i \\ \dot{z}_c
\end{bmatrix}^T \!\!\! P 
\begin{bmatrix}
	z_c-z^*_i \\ \dot{z}_c,
\end{bmatrix} 
$$
where $P$ is defined in \eqref{eq:P}.
Then, \eqref{eq:basic_LMI_matrix} implies that there exists some (small) $\varepsilon > 0$ such that 
$$
\dot{V} \leq -\frac{1}{\gamma} \left(  |W_1 (z_d-h_d(q_i^*))|_2^2  +| W_2\dot{z}_d|_2^2 \right) + \gamma |F_d|_2^2 
$$
as long as $|z_c-z^*_i| + |\dot{z}_c| \leq \varepsilon$. For the initial condition $z_c(0) = z^*_i$, this implies that 
$$
\int_0^t \!\! |W_1 (z_d(\tau)-h_d(q_i^*))|_2^2  + | W_2\dot{z}_d(\tau)|_2^2 d\tau \leq \gamma^2 \! \! \int_0^t  \!\! |F_d(\tau)|_2^2  d\tau 
$$
Thus, if the perturbation $F_d$ is small enough to keep  $|z_c(t)-z^*_i| + |\dot{z}_c(t)| \leq \varepsilon$ for all $t \geq 0$,  
we can extend the interval of integration to $t\to \infty$, arriving at 
$$
\left\| \begin{bmatrix} W_1(z_d-h_d(q_i^*)) \\ W_2 \dot{z_d} \end{bmatrix} \right\|_2^2 \leq \gamma^2 \|F_d\|_2^2 \ . \vspace{-5mm}
$$
\end{proofof}

%% file: content/16-Section4.tex
\section{Examples} \label{examples}

\subsection{2 link end-effector control}

We illustrate our design on the two link manipulator in Figure \ref{fig:twolink}. We take links of length $1$m, each with a mass of $3$Kg. The controller \eqref{eq:control_law_u}, \eqref{eq:control_law_Fc}
corresponds to the one represented in  Figure \ref{fig:twolink}, with virtual spring and dampers aligned to Cartesian axes, pulling the end effector position $z_c$ towards a desired reference $r = [1.0; 1.0]$.

For simplicity, we assume that the disturbing force acts on the end effector as well, that is, $z_d = z_c$. We choose the following weighting matrices, which add a larger weight to vertical displacement/velocity errors.
In a pick and place task, this choice could be motivated by the desire of reaching a tighter control of the vertical position of the robot,
to compensate for uncertainties on the load.  
\begin{align*}
	W_1 &= \begin{bmatrix}
		1.0 & 0.0 \\
		0.0 & 10.0
	\end{bmatrix}
	& W_2 &= \begin{bmatrix}
		0.01 & 0.0 \\
		0.0 & 0.1
	\end{bmatrix}
\end{align*}
We also consider three poses, each requiring the end effector to be in the upper right quadrant. $q_b^*$ corresponds to $r$.
\begin{align*}
	q_a^* &= \begin{bmatrix}
		-1.0 \\ \pi/2 +1.0
	\end{bmatrix},&
	q_b^* &= \begin{bmatrix}
		0.0 \\ \pi/2 
	\end{bmatrix},&
	q_c^* &= \begin{bmatrix}
		1.0 \\ \pi/2 - 1.0
	\end{bmatrix},&
\end{align*}

We construct \textbf{LMI 2} for two different gains: $\gamma=10.0$, and $\gamma=0.1$. The choice of $\gamma$ determines the level of performance, with smaller $\gamma$ corresponding to tighter position control (higher stiffness of the robot). Solving the LMIs with CVX leads to the following controller gains:
\begin{table}[h]
\begin{center}
	\begin{tabular}{cc|cc}
		 \multicolumn{2}{c|}{$\gamma = 10$}& \multicolumn{2}{c}{$\gamma = 0.1$}  \\
		\hline 
		$k = \input{generated_tex/Kp1.tex}$ & $b = \input{generated_tex/Kd1.tex}$
		&
		$k = \input{generated_tex/Kp2.tex}$ & $b = \input{generated_tex/Kd2.tex}$
	\end{tabular} \vspace{-2mm}
\end{center}
\end{table}

The stiffness of the virtual vertical spring is always larger than the stiffness of the horizontal spring, in agreement with the weighting functions $W_1$, and both stiffness becomes larger for $\gamma=0.1$, as expected.
The damping coefficients are larger in the $x$ direction. This can be explained by the greater inertia of the arm in that direction, considering the poses specified in the synthesis. Moving the end-effector in $y$, requires accelerating mostly the second link, whereas moving the end-effector in $x$ requires accelerating both links, with a higher combined inertia.

\begin{figure}[h]
	\centering
	\includegraphics[width=.90\columnwidth]{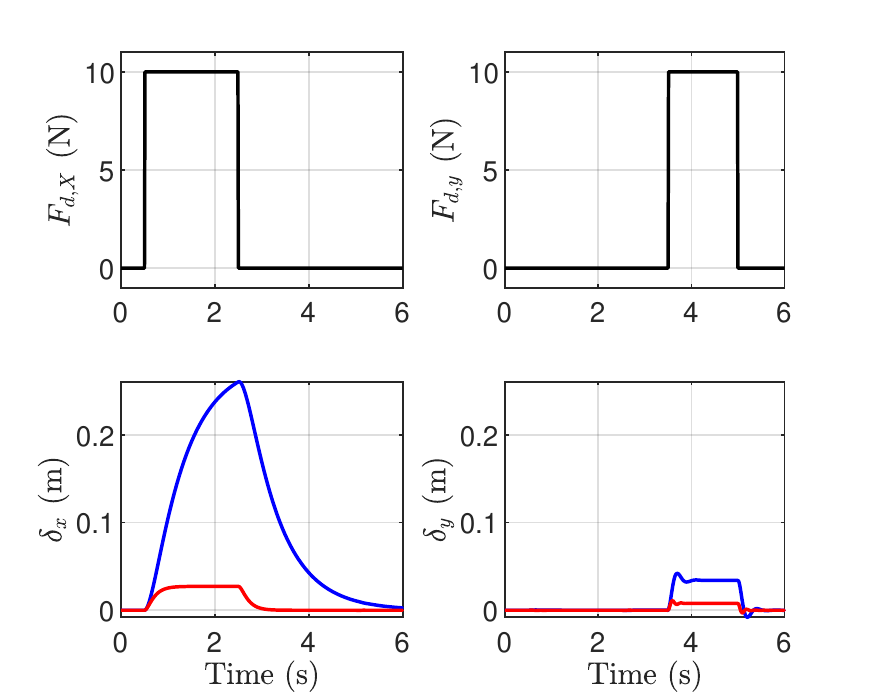} 
	\vspace{-2mm}
	\caption{Disturbance response: {\color{blue}blue - $\gamma=10.0$},  {\color{red}red - $\gamma=0.1$}.}
	\label{fig:Example2LinkDisturbance}
\end{figure}

The control performance is illustrated in Figure \ref{fig:Example2LinkDisturbance}. 
The step disturbance force acts first in the $x$ direction, then in the $y$ direction. 
Larger displacement errors $\delta_x = x - r_x$ and $\delta_y = y - r_y$ correspond to the case $\gamma=10$, as expected. 
In both cases, $\delta_y$ is smaller than $\delta_x$, in agreement with the weights $W_1$ and $W_2$.

\subsection{Control of a redundant arm}
\begin{figure}[h]
	\centering
	\includegraphics[width=.7\columnwidth]{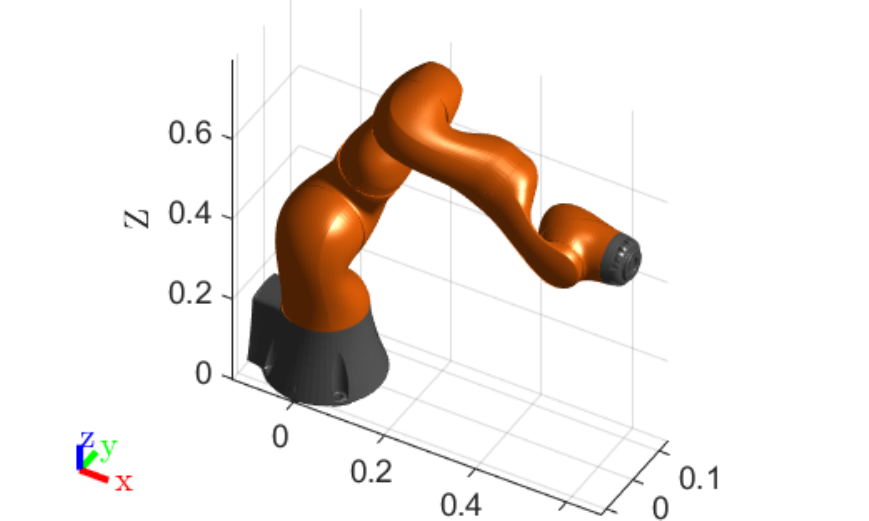} 
	\caption{Pose $q^*$}
	\label{fig:sevenlink}
\end{figure}

We show that the proposed method easily scales to a 7DOF arm, the KUKA LBR iiwa 14 R820 7-axis robot in Figure \ref{fig:sevenlink}, included in the MATLAB Robotics System Toolbox.

We choose a hybrid, 7 dimensional, operation-space comprised of 3 elements. 
The first is the end-effector cartesian position, $z_{ee}(t)$. 
The second is the elbow height, represented by the $z$ coordinates of  link 4, $z_{el}(t)$. 
The final element is the vector of joint angles 5, 6, and 7, $z_q(t)$, related to the final three joints of the arm. 
These angles can be used to set the orientation of the end-effector. Stacking all three vectors together forms $z_c = \begin{bmatrix} z_{ee}^T & z_{el}^T & z_q^T \end{bmatrix}^T$. 

For the disturbing force $F_d$ we consider $z_d = [ \, z_{ee} \ z_q \, ]$, and we choose weighting matrices
$$
	W_1 = 10W_2 = \mathrm{diag}(\begin{bmatrix}
		1 & 1 & 1 & 1 & 1 & 1 
	\end{bmatrix}),
$$
which removes the displacement of the elbow from our performance criterion.

The controller is synthesized around pose $q^*$, shown in Figure \ref{fig:sevenlink}. $q^*$ is not near any extrema of the workspace or singularities. 
The reference position $r$ satisfies $r = z_c^* = h_c(q^*)$, that is, 
\begin{align*}
	q^* &= \input{generated_tex/7linkq}^T \\[5pt]
	r &=\input{generated_tex/7linkr}^T
\end{align*}

Using \textbf{LMI 2}, we derive the controller gains below. These are obtained near $q^*$ for $\gamma = 10$ and $\gamma = 1$, as in the previous example.
Indeed, larger gains corresponds to $\gamma = 1$, as expected.
\begin{table}[h]
\begin{center}
	\begin{tabular}{cc|cc}
		\multicolumn{2}{c|}{$\gamma = 10$}& \multicolumn{2}{c}{$\gamma = 1$}  \\
		\hline
		\!\!$k = \input{generated_tex/4LKp1.tex}$\!\!\!\!\!& \!\!\!\!\!
		$b = \input{generated_tex/4LKd1.tex}$ &
		$k = \input{generated_tex/4LKp2.tex}$ \!\!\!\!\!&\!\!\!\!\!
		$b = \input{generated_tex/4LKd2.tex}$\!\!
	\end{tabular}
\end{center}
\end{table} \vspace{-3mm}

\begin{figure}[h]
	\centering
	\includegraphics[width=.90\columnwidth]{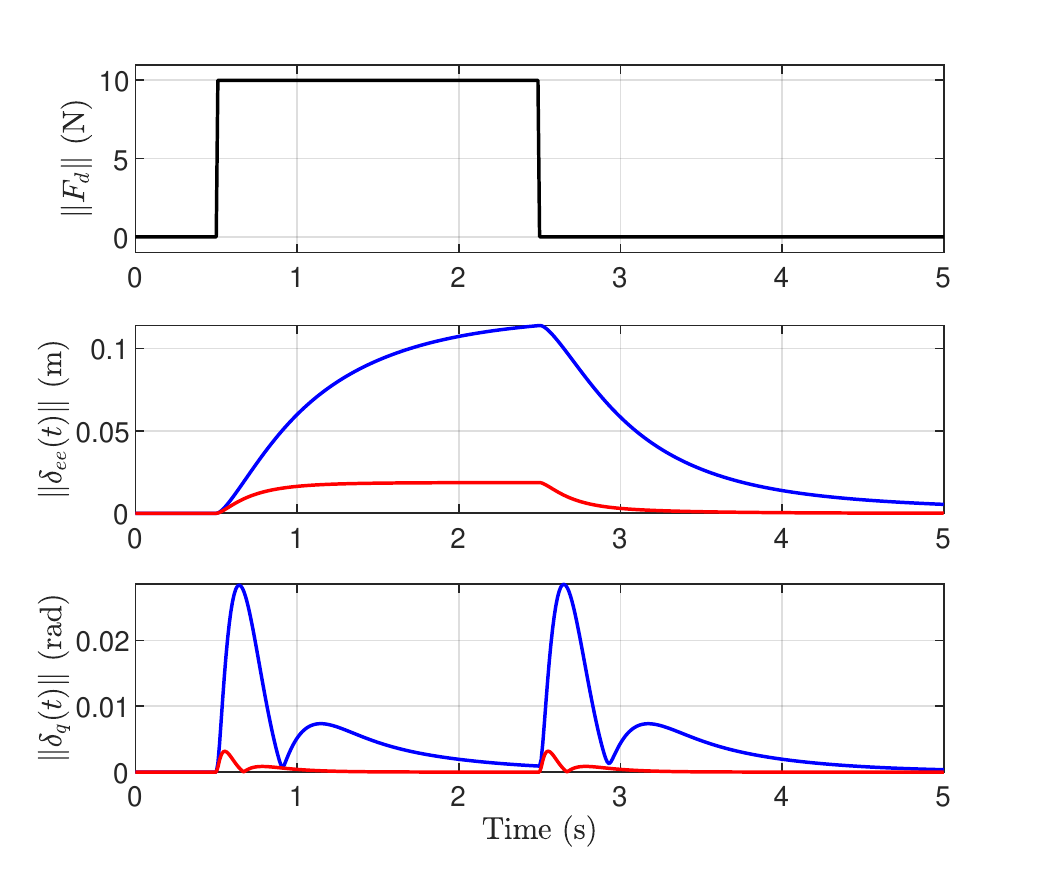} 
	\vspace{-3mm}
	\caption{Disturbance response: {\color{blue}blue - $\gamma=10.0$},  {\color{red}red - $\gamma=1$}.}
	\label{fig:Example7LinkDisturbance}
	\vspace{-4mm}
\end{figure}

The arm is subject to force $F_d = \frac{1}{\sqrt{3}}[\, 10 \ 10\ 10\ 0\ 0 \ 0]^T$. That is, the force acts only on the end effector in direction $\begin{bmatrix}1&1&1\end{bmatrix}$, with magnitude $10$N. 
The norm of the position error at the end effector, 
$\delta_{ee}(t) = z_{ee}(t) - \begin{bmatrix}0.68&0&0.54 \end{bmatrix}^T $ and the norm of the error in joint angles 
$\delta_q(t) = z_q(t) - \begin{bmatrix} 0&-0.5&0 \end{bmatrix}^T$ are displayed in Figure \ref{fig:Example7LinkDisturbance}.
There is a transient error in orientation both at the onset and removal of the disturbance, and a step error in position. Decreasing $\gamma$ increases the synthesized gains and reduces the magnitude of the errors.

%% file: generated_tex/Kp1.tex
\begin{bmatrix}
35.38\\
293.61\\
\end{bmatrix}

%% file: generated_tex/Kd1.tex
\begin{bmatrix}
31.2\\
14.32\\
\end{bmatrix}

%% file: generated_tex/Kp2.tex
\begin{bmatrix}
367.52\\
1267.59\\
\end{bmatrix}

%% file: generated_tex/Kd2.tex
\begin{bmatrix}
82.57\\
20.3\\
\end{bmatrix}

%% file: generated_tex/7linkq.tex
\begin{bmatrix}
0&0.5&0&-1.5708&0&-0.5&0\\
\end{bmatrix}

%% file: generated_tex/7linkr.tex
\begin{bmatrix}
0.68&0&0.54&0.7&0&-0.5&0\\
\end{bmatrix}

%% file: generated_tex/4LKp1.tex
\begin{bmatrix}
316.01\\
50.4\\
179.89\\
1.55\\
2.24\\
22.38\\
0.32\\
\end{bmatrix}

%% file: generated_tex/4LKd1.tex
\begin{bmatrix}
522.8\\
32.5\\
264.57\\
104.17\\
0.35\\
11.48\\
0.04\\
\end{bmatrix}

%% file: generated_tex/4LKp2.tex
\begin{bmatrix}
1746.61\\
326.15\\
1118.97\\
1.2\\
18.15\\
135.98\\
7.4\\
\end{bmatrix}

%% file: generated_tex/4LKd2.tex
\begin{bmatrix}
1034.74\\
76.36\\
596.19\\
77.54\\
1.42\\
26.6\\
0.64\\
\end{bmatrix}

%% file: content/13-Conclusions.tex
\section{Conclusions}

We have shown a procedure to find the gains of an energy shaping and damping injection controller, which optimize a suitable performance metric. Our approach derives controller that guarantees passivity of the controlled robot and local $\mathcal{L}_2$ gain between selected force/displacement variables.

Our LMI-based synthesis requires a diagonal structure of the solution matrices, which severely limit the degrees of freedom of the controller. Future research directions will focus on the relaxation of these sparsity patterns. The goal is to allow for more degrees of freedom while preserving closed-loop passivity. 

One of the reasons to restrict our study to diagonal matrices is that this structural assumption allows for a controller that preserves passivity, as long as we constraint $Q_{12}\prec 0$ in \textbf{LMI 2}. One potential route to generalize this is to leave more freedom on the structure of $Q$ and $L$ in \textbf{LMI 2} while adding explicit LMI conditions for passivity.
Unfortunately, this would lead to conservative results on the gain. Furthermore, enforcing local passivity from $F_d$ to some colocated velocity $\dot{z}_d$, would require that
$P_{12} = 0$ and that $P_{22}$ matches a different expression at each $q_i^*$. This leads to unfeasibility.

The weighting matrices \eqref{eq:weighted_variables} can be extended to weighting transfer matrices, which would apply a nonuniform weighting to the frequency content of $z_d$ and $\dot{z}_d$.
We did not discuss this case for reasons of simplicity and readability of the exposition. This extension would require to endow the linearization in (7) with additional dynamics corresponding to a minimal realization of $W_1(s)$ and $W_2(s)$. We would also need to enforce zero sub-blocks constraints in the matrices $Q$ and $L$ of \textbf{LMI 2}, to obtain a state feedback of the right dimension.

Finally, in this paper we have focused on the design of $N$ stiffness and $N$ damping parameters, where $N$ are the degrees of freedom of the robot. As future research we will explore cases where the number of control parameters is larger than the degrees of freedom of the robot. In the interpretation of virtual springs and dampers, this corresponds to the case that a large number of springs and damper is employed to shape the energy of the robot. This could be approached, for example, using \textbf{LMI 2} repeatedly on different subsets of $2N$ parameters (each time keeping the other parameters constant).